\documentclass[12pt,fleqn]{article}
\begin{document}
\bigskip
\begin{center}
{\sffamily\bfseries\LARGE
    On the monotonization of the training set
}\\[1em]
{\bfseries
Rustem S. Takhanov
}\\[1em]

\end{center}
\thispagestyle{empty}

\begin{abstract}
We consider the problem of minimal correction of the training set
to make it consistent with monotonic constraints. This problem
arises during analysis of data sets via techniques that require
monotone data. We show that this problem is NP-hard in general and
is equivalent to finding a maximal independent set in special
orgraphs. Practically important cases of that problem considered
in detail. These are the cases when a partial order given on the
replies set is a total order or has a dimension 2. We show that
the second case can be reduced to maximization of a quadratic
convex function on a convex set. For this case we construct an
approximate polynomial algorithm based on convex optimization.

\end{abstract}

\section{Introduction}
Requirements to a classifying rule in supervised learning problems
consist of two parts. The first part is induced by a set of
precedents, called the training set. Each element in the training
set is a pair of "object--reply" type. A classifying rule which is
a mapping from objects set to the replies set should map objects
from the training set pairs to the consistent replies. And the
second part of requirements express our common knowledge of a
classifying rule. One of the popular types of such requirements is
the monotonicity which is considered in that paper. In some cases
these two parts of requirements can not be satisfied both and then
we have a problem of a minimal correction of the training set. Let
us see what that problem is.

Suppose the sets $X,Y$ are given and on this sets we have partial
orders $ \ge ^X , \ge ^Y $ consistently. We assume more that the
partial order $ \ge ^Y $ is a lattice. For any given mapping $o:X'
\to Y$ where $X' \subseteq X,\left| {X'} \right| < \infty $ we
pose a problem of finding a function $f:X \to Y$ which is monotone
due to partial orders $ \ge ^X , \ge ^Y $ and minimizes the
following functional: $Er_o \left( f \right) = \left| {\left\{
{x|f\left( x \right) \ne o\left( x \right)} \right\} \cap X'}
\right|$.

Let us denote the set of monotonic functions from $X$ to $Y$ by
$M\left( { \ge ^X , \ge ^Y } \right)$. Then for a given mapping
$o:X' \to Y$ our task is the following:
\[
 Er_o
\left( f \right)\rightarrow \mathop{\min}\limits_{f\in M\left( {
\ge ^X , \ge ^Y } \right)}
\]

Every mapping $f':X' \to Y$ which is monotone on the subset $X'
\subseteq X$ can be extended to the mapping monotone on the whole
set $X$ because $\left( {Y, \ge ^Y } \right)$ is a lattice.
Actually on every finite subset of the lattice $\left( {Y, \ge ^Y
} \right)$ the operation $\sup $ is defined and the function
$f\left( x \right) = \sup \left\{ {f'\left( {x'} \right)|x'\in X',
x' \le ^X x} \right\}$ is both monotone and satisfies $f\left( x
\right) = f'\left( x \right),x \in X'$. From this we see that in
the posed problem we can imply that $X' = X$. From the above said
we conclude more that this problem is equivalent to finding a
maximal subset $X'' \subseteq X'$ such that the function $o$
restricted on the subset $X''$ is monotone.

So let us consider the following generalization of our problem
which we will call MaxCMS(Maximal Consistent with Monotonicity
Set).

{\bf MaxCMS.} The finite sets $B_n ,B_m $ where $B_r  = \left\{
{1,...,r} \right\}$ are given; on each of them partial orders $
\ge ^1 , \ge ^2 $ are defined consistently and the function
$\varphi :B_n \to B_m $ is given. Then every element $i \in B_n $
is assigned by a positive integer weight $w_i $. Our task is to
find a maximal by weight subset $B \subseteq B_n $ such that the
function $\varphi $ restricted on $B$ is monotone i.e. $\forall
i,j \in B\left[ {i \ge ^1 j \to \varphi \left( i \right) \ge ^2
\varphi \left( j \right)} \right]$ .

{\bf Definition 1.} The set $B \subseteq B_n $ is called
acceptable iff the function $\varphi $ restricted on $B$ is
monotone.

{\bf Definition 2.} A set which  is acceptable and maximal by
weight is denoted by $MaxCMS\left( { \ge ^1 , \ge ^2 ,\varphi ,w}
\right)$(in some cases we use this notation to mean the weight of
this set).

In the remainder of the paper we will consider that problem.

\section{Training set monotonization and maximal independent sets}
In this section we will show that MaxCMS is equivalent to finding
a maximal independent set(or minimal vertex cover) in special
orgraphs.

{\bf Definition 3.} Let $G = \left( {V,E} \right)$ be an orgraph
and every vertex $v$ of an orgraph has a positive integer weight
$w_v $. A set of vertexes is called independent iff every pair of
its elements is not connected by an edge. The maximal by weight
independent set is denoted by $IS\left( {G,w} \right)$ (in some
cases we use this notation to mean the weight of this set).

As well-known, the supplement of independent set is vertex cover.

Let us define the following partial preorder on $B_n $ (recall,
that it means transitive and reflexive binary predicate):
 \[
 i \succ j \Leftrightarrow \varphi \left( i \right) \ge ^2 \varphi \left( j
 \right).
 \]

Consider the orgraph $G = \left( {V,E} \right)$ with $V = B_n$ and
$E = \left\{ {\left( {i,j} \right)|i \ge ^1 j,\varphi \left( i
\right)\not  \ge ^2 \varphi \left( j \right)} \right\}$. The
orgraph $G$ can alternatively be defined through the following
equalities: $V = B_n $ and $E =  \ge ^1  \cap \overline  \succ$
where $\overline\succ $ is a supplement of the binary predicate.

{\bf Definition 4.} An orgraph which has the edge set represented
as a intersection of a partial order and a supplement of a partial
preorder is called special.

{\bf Theorem 1.} The maximal acceptable set is equal to the
maximal independent set of the special orgraph $G$, i.e.
$MaxCMS\left( { \ge ^1 , \ge ^2 ,\varphi ,w} \right) = IS\left(
{G,w} \right)$.

{\bf Proof.} Any independent set $B$ of the orgraph $G$ satisfies
the condition: if $i,j \in B$ and $i \ge ^1 j$ then $\varphi
\left( i \right) \ge ^2 \varphi \left( j \right) $, i.e. the
function $\varphi $ restricted on $B$ is monotone. The inverted
statement is correct also: if restriction of $\varphi $ on $B$ is
monotone then $B$ is an independent set in $G$. From this we
obtain the proposition of the theorem.

{\bf  Theorem 2.} Let the special orgraph $G'$ be defined by the
vertex set $V' = B_n $ with weights $w'_i $ and the edge set $E' =
\ge ' \cap \overline { \succ '}$; both $\ge '$ and $\overline {
\succ '}$ are given (i.e. the edge set $E'$ need not be
decomposed). Then the problem of finding maximal independent set
in such an orgraph polynomially reducible to MaxCMS.

{\bf Proof.} Let us divide the set $V'$ on the equivalence classes
due to predicate $x \sim y \Leftrightarrow x \succ' y\& y \succ'
x$. Then we can naturally define the corresponding mapping
$\varphi ':V \to V/\sim $. On the factor-set $V/\sim $ it is
induced the partial order $\overline x \ge ''\overline y
\Leftrightarrow x \succ' y$. It is easy to see that $IS\left(
{G',w'} \right) = MaxCMS\left( { \ge ', \ge '',\varphi ',w'}
\right)$. The reduction is done in $O\left(n^{2}\right)$ steps.

\section{NP-hardness of MaxCMS.}
In the previous chapter it was shown that MaxCMS is equivalent to
finding maximal independent set(or minimal vertex cover) in
special orgraphs. The problem of finding an acceptable set of
cardinality more than $C$ is denoted by CMS. Obviously, it is in
NP.

{\bf Theorem 3.} CMS is NP-complete.

{\bf Proof.} Let us reduce CMS to 3-SAT using the trick from
$\cite{garyjohn}$.

Let 3-CNF be given with $U = \left\{ {u_1 ,...,u_n } \right\}$
being the set of variables used in it. Let $C = \left\{ {c_1
,...,c_m } \right\}$ be the set of clauses such that each clause
consists of 3 literals that differ by their variables (literal is
symbol $u_i $ or $\overline {u_i } $). For every clause we order
literals that belong to it. Then the fact of meeting the literal
$l$ on the $s$-th place in the clause $c_r $ is denoted by $lc_r^s
$. Let us consider the orgraph such that its vertex set is a union
of all literals and threefold copies of clauses $V = \left\{ {u_1
,\overline {u_1 } ...,u_n ,\overline {u_n } } \right\} \cup
\left\{ {c_1^1 ,c_1^2 ,c_1^3 ,...,c_m^1 ,c_m^2 ,c_m^3 } \right\}$.
Let us define the edge set being equal to $E = E_1 \cup E_2 $
where $E_1 = \left\{ {\left( {u_i ,\overline {u_i } } \right)}
\right\}_{i = 1}^n  \cup \left\{ {\left( {u_k ,c_m^l } \right)|u_k
c_m^l } \right\} \cup \left\{ {\left( {c_m^l ,\overline {u_k } }
\right)|\overline {u_k } c_m^l } \right\}$ and $E_2  = \left\{
{\left( {c_j^1 ,c_j^2 } \right),\left( {c_j^2 ,c_j^3 }
\right),\left( {c_j^1 ,c_j^3 } \right)} \right\}_{i = 1}^m $
(later we will need this division of the edge set on 2 subsets).

A vertex cover of the orgraph $G = \left( {V,E} \right)$ of the
cardinality $n + 2m$ exists iff the original 3-CNF is satisfiable.
Actually, one from every pair of vertexes $u_i ,\overline {u_i } $
and two from every triple $c_j^1 ,c_j^2 ,c_j^3 $ should fall into
the vertex cover, because they are pairwise connected. And so, the
cardinality of a vertex cover is not less than $n + 2m$.

Suppose the vertex cover of the required cardinality exists. If
the literal $u_i $ is in it we define $u_i  = true$, otherwise
$u_i = false$. All variables should be initialized in this manner,
because from the above said it is clear that $u_i $ or $\overline
{u_i } $ is in the cover excluding both of them. Then this
assignment, as easily seen, satisfies the original 3-CNF. This
reasoning can be inverted and we obtain that the existence a
satisfying assignment is equivalent to the existence of a vertex
cover of the cardinality $n + 2m$.

Now let us consider the orgraph $G' = \left( {V,E^* \backslash E}
\right)$ where $E^* $ is a transitive closure of $E$. Suppose that
the edge set of $G'$ is transitive. Then defining $ \ge  = E^* $
and $ \succ  = E^* \backslash E$ we obtain that $ \ge  \cap
\overline \succ = E$. This means that our problem is reduced to
finding the minimal vertex cover, and consequently, the maximal
independent set of the special orgraph $G = \left( {V,E} \right)$,
which is by theorem 2 is equivalent to MaxCMS, or CMS when $C = 2n
+ 3m - (n + 2m) = n + m$.

Let us show that the edge set of $G'$ is transitive. As $E^*$ is
transitive, $E^* \backslash E$ is not transitive only if there
exists such $\left( {u,v} \right),\left( {v,t} \right) \in E^*
\backslash E$ that $\left( {u,t} \right) \in E$. Let $\left( {u,t}
\right) \in \left\{ {\left( {u_i ,\overline {u_i } } \right)}
\right\}_{i = 1}^n $. It is easy to see that any path in the $G$
starting with $u_i $ can not end with literal $\overline {u_i } $,
because otherwise there should exist a clause that contains both
$u_i $ and $\overline {u_i } $. Let us now consider the case when
$\left( {u,t} \right) \in \left\{ {\left( {c_j^1 ,c_j^2 }
\right),\left( {c_j^2 ,c_j^3 } \right),\left( {c_j^1 ,c_j^3 }
\right)} \right\}$. In that case the path starting from
$c_j^\alpha $ and finishing in $c_j^\beta $ can not contain an
element which does not belong to $\left\{ {c_j^1 ,c_j^2 ,c_j^3 }
\right\}$. Consequently, $\left( {u,v} \right),\left( {v,t}
\right) \in E$, which contradicts to $\left( {u,v} \right),\left(
{v,t} \right) \in E^* \backslash E$. And the last case is when
$\left( {u,t} \right) \in \left\{ {\left( {u_k ,c_m^l }
\right)|u_k c_m^l } \right\}$. But every path in orgraph $G$ which
starts in $u$ and finishes in $t$ is equal to edge $\left( {u,t}
\right)$, and this means $\left( {u ,v } \right)\not \in E^*
\backslash E$. In the same manner the case $\left( {u,t} \right)
\in \left\{ {\left( {c_m^l, \overline {u_k}} \right)| \overline
{u_k}c_m^l } \right\}$ is considered. So, the set $E^* \backslash
E$ is transitive and the reduction of 3-CNF to CMS is done.

\section{1-MaxCMS}
Any partial order on a finite set can be represented as
intersection of total orders.

{\bf Definition.} Let the partial order $ \ge $ be given on the
set $M$. The minimal number $d$ such that $ \ge $ is an
intersection of total orders $ \ge _1 ,..., \ge _d $, i.e. $ \ge =
\ge _1 \cap ... \cap  \ge _d $, is called the dimension of $ \ge
$.

Consider MaxCMS with input $\left( { \ge ^1 , \ge ^2 ,\varphi ,w}
\right)$ in case when the dimension of $ \ge ^2 $ is equal to $d$.
In that case $ \ge ^2  =  \ge _1  \cap ... \cap  \ge _d $. The
consistent special orgraph $G = \left( {V,E} \right)$ satisfies:
$V = B_n $ and $E =  \ge ^1  \cap \overline  \succ$ where
 $i \succ j \Leftrightarrow i \succ _1 j\& ...\& i \succ
_d j$ and $i \succ _s j \Leftrightarrow \varphi \left( i \right)
\ge _s \varphi \left( j \right)$. And then,
\[
E =  \ge ^1  \cap \overline { \succ _1  \cap ... \cap  \succ _d }
=  \ge ^1  \cap \left( {\overline { \succ _1 } \cup ... \cup
\overline { \succ _d } } \right) =\left( { \ge ^1 \cap \overline {
\succ _1 } } \right) \cup ... \cup \left( { \ge ^1  \cap \overline
{ \succ _d } } \right).
\]

As each predicate $ \overline {\succ _s}$ is transitive, $E$ is a
union of $d$ transitive predicates.

{\bf Definition.} The problem MaxCMS with input $\left( { \ge ^1 ,
\ge ^2 ,\varphi ,w} \right)$ for case when the dimension of $ \ge
^2 $ is equal to $d$ is called $d$-MaxCMS.

In fact, the above mentioned showed that

{\bf  Theorem 4.} The problem $d$-MaxCMS is reduced to finding the
maximal independent set in the orgraph $G = \left( {V,E} \right)$
where $ E =  \succ ^1  \cup ... \cup  \succ ^d$ and predicates
$\succ ^s$ are transitive and there are no cycles in $G$.

From the theorem 4 we see that 1-MaxCMS is reduced to finding the
maximal independent set in the circuit-free orgraph $G = \left(
{V,E} \right)$ that has the edge set satisfying the following
transitivity rule: if $\left( {u,v} \right),\left( {v,t} \right)
\in E$ then $\left( {u,t} \right) \in E$. This problem is
polynomially tractable because the graph that can be obtained from
$G$ by transformation of oriented edges to non-oriented is a
comparability graph of some partial order which is known to be
perfect. We will adduce one of the proofs of the tractability due
to\cite{mohring}.

{\bf Theorem 5.} 1-MaxCMS is polynomially tractable.

{\bf Proof.} Defining $x \triangleright y \Leftrightarrow \left(
{x,y} \right) \in E$, the orgraph can be seen as partially ordered
set $\left( {V, \triangleright } \right)$. The algorithm solves
the problem via reducing it to the task of minimizing a flow in
some circuit-free network. Let us denote the sets of minimal and
maximal elements of $\left( {V, \triangleright } \right)$ by $\min
G$ and $\max G$ consistently. For every vertex $v \in V$ of the
orgraph $G$ we introduce 2 copies $v^ +  ,v^ -  $. And then we
define $V' = \left\{ {v^ + ,v^ -  } \right\}_{v \in V}  \cup
\left\{ {s,t} \right\}$ and $E' = \left\{ {\left( {v^ +  ,v^ -  }
\right)} \right\}_{v \in V} \cup \left\{ {\left( {x^ -  ,y^ +  }
\right)|\left( {x,y} \right) \in E} \right\} \cup \left\{ {\left(
{s,a^ +  } \right)|a \in \min G} \right\} \cup \left\{ {\left( {b^
-  ,t} \right)|b \in \max G} \right\}$. We obtained the orgraph
$G' = \left( {V',E'} \right)$. The minimal flow through the edge
$\left( {v^ + ,v^ -  } \right)$ is defined to be equal to the
corresponding weights $w_v $, and for other edges it equals 0. The
maximal flow through every edge is $\infty $. It is easy to see
that for every edge $e \in E'$ of the orgraph $G'$ we can find a
path from $s$ to $t$ that goes through $e$. It is well-known that
under that condition we can apply the min flow-max cut theorem.

The minimal flow of given network, that can be obtained via
modified Ford-Fulkerson algorithm(common algorithm finds maximal
flow), corresponds to the maximal W-cut(common algorithm finds
minimal cut), where by the weight of a cut we mean the following
expression:
\[
\sum\limits_{\left( {u,v} \right) \in E,u \in S,v \in \overline S
} {c_{\min } \left( e \right)}  - \sum\limits_{\left( {u,v}
\right) \in E,v \in S,u \in \overline S } {c_{\max } \left( e
\right)}.
\]
Note that the weight of a cut is defined differently from the sum
of weights between parts of a cut and that is why we call the
problem maximal W-cut. Consider any cut $V' = S \cup \overline S $
where $s \in S,t \in \overline S $ with the weight different from
$ - \infty $. Since maximal flow through edges is $\infty $, for
every edge $\left( {u,v} \right) \in E'$ there can not be $v \in
S,u \in \overline S $. And edges $\left( {u,v} \right) \in E'$ for
$u \in S,v \in \overline S $ can make a contribution to the weight
of a cut only when $u = r^ + ,v = r^ - $. Let us denote $R =
\left\{ {r|r^ + \in S,r^ -   \in \overline S } \right\}$.
Obviously, the elements of $R$ constitute an independent set in
$G$ and the weight of a cut is exactly equal to the weight of the
set. The conversion of the statement is also correct, i.e. every
independent set $R$ of $G$ correspond to the cut $S = \left\{ {u^
+ ,u^ -  |u \notin R\& \exists r \in R\left[ {r \triangleright u}
\right]} \right\} \cup \left\{ {r^ +  |r \in R} \right\} \cup
\left\{ s \right\}$, the weight of a cut being equal to the weight
of $R$. From this it is clear that the result of an algorithm will
be the maximal cut that correspond to the maximal independent set
in $G$. The theorem proved.

The task of finding the minimal flow can be written in the LP
form:
\[
\begin{array}{*{20}c}
x\left( \Gamma  \right) \ge 0,\Gamma  \in G\left( s,t \right) \\
\sum\limits_{\Gamma  \in G\left( v \right)} {x\left( \Gamma  \right)}  \ge w_v   \\
\sum\limits_{\Gamma  \in G\left( {s,t} \right)} {x\left( \Gamma  \right)}  \to \min   \\
\end{array}
\]
where $G\left( {s,t} \right)$ is a set of paths in orgraph $G' =
\left( {V',E'} \right)$ from $s$ to $t$, and $G\left( v \right)
\subset G\left( {s,t} \right)$ is a set of paths going through the
edge $\left( {v^ + ,v^ - } \right)$. In the dual form:
 \[
 \begin{array}{*{20}c}
   {y\left( v \right) \ge 0,v \in V}  \\
   {\sum\limits_{\left( {v^ +  ,v^ -  } \right) \in \Gamma } {y\left( v \right)}  \le 1,\Gamma  \in G\left( {s,t} \right)}  \\
   {\sum\limits_{v \in V} {w_v y\left( v \right) \to \max } }  \\
\end{array}
\]

From the above stated we conclude that the dual problem always has
a boolean solution. Polyhedron of the dual problem is denoted by
$\Pi \left( G \right)$.

\section{2-MaxCMS}
Now we will consider the problem 2-MaxCMS. This problem arise when
a partial order on the replies set is not total, but, for example,
has a tree structure. As we know, it can be reduced to finding the
maximal independent set in the circuit-free orgraph $G = \left(
{V,E} \right)$ where $E = \succ ^1 \cup \succ ^2$ and the
predicates $\succ ^s$ are transitive. From now on we will consider
just that problem.

Note that edges of the circuit-free orgraph from theorem 3 are
also divided on 2 sets $E_1$ and $E_2 $, both of them being
transitive. From this we conclude that the problem is NP-hard.

Consider 2 orgraphs: $G_1  = \left( {V, \succ ^1 } \right)$ and
$G_2 = \left( {V, \succ ^2 } \right)$. Note that the maximal
independent set of the orgraph $G = \left( {V,E} \right)$ is also
an independent set in both $G_1 $ and $G_2 $. Then the following
theorem is obvious.

{\bf Theorem 6. } The set of solutions to the following quadratic
programming problem
\[
\begin{array}{*{20}c}
   {\overline x  \in \Pi \left( {G_1 } \right)}  \\
   {\overline y  \in \Pi \left( {G_2 } \right)}  \\
   {\psi \left( {\overline x ,\overline y } \right) = \sum\limits_{v \in V} {w_v x_v y_v }  \to \max }  \\
\end{array}
\]
contains such boolean $\overline x ^* ,\overline y ^*$ that
$\left\{ {v|x_v^* y_v^* = 1} \right\}$ is the maximal independent
set in $G$.

{\bf Proof.} With fixed $\overline x$(fixed $\overline y$) the
maximum of $\sum\limits_{v \in V} {w_v x_v y_v }$ is reached on
some boolean $\overline y$(boolean $\overline x$). It means that
the maximum by both vectors can be achieved with boolean values of
components.

{\bf Theorem 7.} The following is true
\[
\mathop {\max }\limits_{\overline x  \in \Pi \left( {G_1 }
\right),\overline y  \in \Pi \left( {G_2 } \right)} \psi \left(
{\overline x ,\overline y } \right) = \mathop {\max
}\limits_{\overline x  \in \Pi \left( {G_1 } \right),\overline y
\in \Pi \left( {G_2 } \right)} \gamma \left( {\overline x
,\overline y } \right),
\]
where
\[
\gamma \left( {\overline x ,\overline y } \right) =
\frac{1}{2}\sum\limits_{v \in V} {w_v \left( {x_v  + y_v }
\right)^2 - w_v \left( {x_v  + y_v } \right)}
\]

{\bf Proof.}
\[
\mathop {\max }\limits_{\overline x  \in \Pi \left( {G_1 }
\right),\overline y  \in \Pi \left( {G_2 } \right)} \sum\limits_{v
\in V} {w_v x_v y_v } = \mathop {\max }\limits_{\overline x \in
\Pi \left( {G_1 } \right),\overline y \in \Pi \left( {G_2 }
\right)} \frac{1}{2}\sum\limits_{v \in V} {w_v \left( {x_v  + y_v
} \right)^2 - w_v \left( {x_v^2  + y_v^2 } \right)} \ge
\]
\[
\ge\mathop {\max }\limits_{\overline x  \in \Pi \left( {G_1 }
\right),\overline y  \in \Pi \left( {G_2 } \right)}
\frac{1}{2}\sum\limits_{v \in V} {w_v \left( {x_v  + y_v }
\right)^2  - w_v \left( {x_v  + y_v } \right)}
\]

Since maximum of the left part of inequality is achieved on
boolean vectors, it is clear that the equality holds. Taking into
account that the functional $\gamma \left( {\overline x ,\overline
y } \right)$ is convex, we see that the problem was reduced to the
maximization of a convex quadratic function on a convex set.

Consider the functional
\[
\varphi \left( {\overline x ,\overline y } \right) =  -
\frac{1}{2}\sum\limits_{v \in V} {w_v \left( {x_v  - y_v }
\right)^2  - w_v \left( {x_v  + y_v } \right)}
\]

{\bf Theorem 8. } The following is true
\[
\mathop {\max }\limits_{\overline x  \in \Pi \left( {G_1 }
\right),\overline y  \in \Pi \left( {G_2 } \right)} \varphi \left(
{\overline x ,\overline y } \right) \ge \mathop {\max
}\limits_{\overline x  \in \Pi \left( {G_1 } \right),\overline y
\in \Pi \left( {G_2 } \right)} \psi \left( {\overline x ,\overline
y } \right),
\]
the values of $\varphi \left( {\overline x ,\overline y } \right)$
and $\psi \left( {\overline x ,\overline y } \right)$ being equal
on the boolean vectors of the polyhedron $\overline x \in \Pi
\left( {G_1 } \right),\overline y \in \Pi \left( {G_2 } \right)$ .

{\bf Proof. } The verification of the second statement is obvious.
The first follows it, because the maximum of the right part by
theorem 6 can be achieved on boolean vectors.

Consider the following optimization task:
\[
\begin{array}{*{20}c}
   {\overline x  \in \Pi \left( {G_1 } \right)}  \\
   {\overline y  \in \Pi \left( {G_2 } \right)}  \\
   {\varphi \left( {\overline x ,\overline y } \right) \to \max }  \\
\end{array}
\]
Let us call it as the convex task.

{\bf Definition. } The pair $\overline x ^*  \in \Pi \left( {G_1 }
\right),\overline y ^* \in \Pi \left( {G_2 } \right)$ such that
$\mathop {\max }\limits_{\overline x \in \Pi \left( {G_1 }
\right),\overline y \in \Pi \left( {G_2 } \right)} \varphi \left(
{\overline x ,\overline y } \right) - \varphi \left( {\overline x
^* ,\overline y ^* } \right) \le \varepsilon$ is called
$\varepsilon $-solution of the convex task.

{\bf Theorem 9.} For every $\varepsilon $ the convex task can be
$\varepsilon $-solved in polynomial time. The length of an input
is a sum of the lengths of descriptions of $G_1 = \left( {V, \succ
^1 } \right)$, $G_2 = \left( {V, \succ ^2 } \right)$ and integer
weights $w_v $. And obtained $\varepsilon $-solution $\left(
{\overline x ^* ,\overline y ^* } \right)$ satisfies $\left|
{x_i^{*} - y_i^{*} } \right| \le \frac{1}{2}$.

{\bf Lemma.} The pair $\overline {\xi ^{opt} }  = \left(
{\overline {x^{opt} } ,\overline {y^{opt} } } \right) = \arg
\mathop {\max }\limits_{\left( {\overline x ,\overline y } \right)
\in \Pi \left( {G_1 } \right) \times \Pi \left( {G_2 } \right)}
\varphi \left( {\overline x ,\overline y } \right)$ satisfies
$\left| {x_i^{opt}  - y_i^{opt} } \right| \le \frac{1}{2}$.

{\bf Proof of lemma.} Quadratic functional $\varphi \left(
{\overline x ,\overline y } \right)$ is not bounded in $R^{2n} $
and its maximum on the set $\Pi \left( {G_1 } \right) \times \Pi
\left( {G_2 } \right)$ is located on the borders of polyhedron.
Let $\overline {a_1 } ^{\rm T} \overline \xi   \le b_1 $ , ... ,
$\overline {a_s } ^{\rm T} \overline \xi   \le b_s $ be those
inequalities from the definition of polyhedron that turn into
equalities. From the optimality of $\left( {\overline {x^{opt} }
,\overline {y^{opt} } } \right)$ it is clear that the cone
$\left\{ {\overline \xi  |\overline {a_1 } ^{\rm T} \overline \xi
\le 0} \right\} \cap ... \cap \left\{ {\overline \xi |\overline
{a_s } ^{\rm T} \overline \xi \le 0} \right\} \cap \left\{
{\overline \xi |\nabla _{\overline {\xi ^{opt} } } \varphi \left(
{\overline {\xi ^{opt} } } \right)^{\rm T} \overline \xi > 0}
\right\} = \emptyset $. And then, from theorem of
Farkas-Minkovski, we conclude that $\varphi \left( {\overline {\xi
^{opt} } } \right)$ can be expanded on positive combination of
vectors $\overline {a_1 } ,...,\overline {a_s } $. But taking into
account that components of those vectors are positive we obtain
that $\nabla _{\overline {\xi ^{opt} } } \varphi \left( {\overline
{\xi ^{opt} } } \right) = \| w_1(x_1^{opt}  - y_1^{opt}  +
\frac{1}{2}),w_1(y_1^{opt}  - x_1^{opt}  + \frac{1}{2})$, ... ,
$w_n(x_n^{opt} - y_n^{opt}  + \frac{1}{2}),w_n(y_n^{opt}  -
x_n^{opt} + \frac{1}{2}) \|^{\rm T}  \ge \overline 0$. Lemma
proved.

{\bf Proof of theorem. } Since the function $\varphi \left(
{\overline x ,\overline y } \right)$ is concave, the set of pairs
\[
\begin{array}{*{20}c}
   {\overline x  \in \Pi \left( {G_1 } \right)}  \\
   {\overline y  \in \Pi \left( {G_2 } \right)}  \\
   {\varphi \left( {\overline x ,\overline y } \right) \ge c}  \\
   {-\frac{1}{2} \le x_i  - y_i  \le \frac{1}{2}}, i = \overline{1,n} \\
\end{array}
\]
is convex.

Note that for every given vector pair $\overline x ^\prime
,\overline y ^\prime $ the task of defining whether it belongs to
the set $\Pi \left( {G_1 } \right) \times \Pi \left( {G_2 }
\right)$ or not can be solved in polynomial time. Actually, by
Floid-Warshall algorithm we can find the longest path from $s$ to
$t$ in orgraphs  $G_1$ and $G_2$ in polynomial time, where by
length of a path we mean a sum of weights of vertexes on the path.
Comparing the results with 1 we see that if they are less than 1
then $\overline x ^\prime ,\overline y ^\prime \in \Pi \left( {G_1
} \right) \times \Pi \left( {G_2 } \right)$. Besides, if
$\overline x ^\prime ,\overline y ^\prime \notin \Pi \left( {G_1 }
\right) \times \Pi \left( {G_2 } \right)$ then the path which
length is greater than 1 will give us a violated inequality in the
definition of the polyhedron $\Pi \left( {G_1 } \right) \times \Pi
\left( {G_2 } \right)$.

And for given $\overline x ^\prime ,\overline y ^\prime $, the
satisfaction of conditions $\varphi \left( {\overline x^\prime
,\overline y^\prime } \right) \ge c$, and in negative case, the
separating hyperplane for the pair $\overline x^\prime ,\overline
y^\prime$ and the set $\left\{ \left( {\overline x ,\overline y }
\right)|\varphi \left( {\overline x ,\overline y } \right) \ge c +
\varepsilon \right\}$ can be found in polynomial time.

Actually,
\[
\begin{array}{*{20}c}
\{{\left( {\overline x ,\overline y } \right) \in \Pi \left( {G_1
} \right) \times \Pi \left( {G_2 } \right)|\left( {\nabla
_{\overline {x'} } \varphi \left( {\overline {x'} ,\overline {y'}
} \right),\overline x  - \overline {x'} } \right) +\left({\nabla
_{\overline {y'} } \varphi \left( {\overline {x'} ,\overline {y'}
} \right),\overline y  - \overline {y'} } \right)
\ge \varepsilon} \} \supseteq  \\
\supseteq \left\{ {\left( {\overline x ,\overline y } \right) \in
\Pi \left( {G_1 } \right) \times \Pi \left( {G_2 } \right)|\varphi
\left( {\overline x ,\overline y } \right) \ge c + \varepsilon} \right\}\\
\end{array}
\]
This can be seen from the following inequalities for concave
quadratic function $\varphi$ and points $\left( {\overline {x}
,\overline {y} } \right)$, $\left( {\overline {x'} ,\overline {y'}
} \right)$ such that $\varphi \left( {\overline {x} ,\overline {y}
} \right) \ge c + \varepsilon$ and $\varphi \left( {\overline {x'}
,\overline {y'} } \right) \le c$: $\varepsilon \le \varphi \left(
{\overline {x} ,\overline {y} } \right) - \varphi \left(
{\overline {x'} ,\overline {y'} } \right) \le \left(\nabla
_{\overline {x'} } \varphi \left(\overline {x'} ,\overline {y'}
\right),\overline x  - \overline {x'}\right) +\left({\nabla
_{\overline {y'} } \varphi \left( {\overline {x'} ,\overline {y'}
} \right),\overline y  - \overline {y'} } \right)$.

Then rounding each component of vectors $\nabla _{\overline {x'} }
\varphi \left( {\overline {x'} ,\overline {y'} } \right)$ and
$\nabla _{\overline {y'} } \varphi \left( {\overline {x'}
,\overline {y'} } \right)$ to the first $2\left( {\log n + \left|
{\log \varepsilon } \right| + 1} \right)$ symbols in binary
representation and denoting them as $c_x $ and $c_y $, will give
us the separating hyperplane
\[
\left\{ {\left( {\overline x ,\overline y } \right)|\left( {c_x
,\overline x  - \overline {x'} } \right) + \left( {c_y ,\overline
y  - \overline {y'} } \right) \ge \frac{{\varepsilon}}{2}}
\right\}.
\]

According to\cite{lovasz}, in this case to find a pair
 $\overline x ^\prime  ,\overline y ^\prime$ that satisfies:
\[
\begin{array}{*{20}c}
   {\overline x^\prime  \in \Pi \left( {G_1 } \right)}  \\
   {\overline y^\prime  \in \Pi \left( {G_2 } \right)}  \\
   {\varphi \left( {\overline x^\prime ,\overline y^\prime } \right) \ge c}  \\
   {-\frac{1}{2} \le x_i^\prime  - y_i^\prime  \le \frac{1}{2}}, i = \overline{1,n} \\
\end{array}
\]
can be done in polynomial time, or it will be shown that
\[
\begin{array}{*{20}c}
\{ \left( {\overline x ,\overline y } \right)|\overline x \in \Pi
\left( {G_1 } \right),\overline y  \in \Pi \left( {G_2 }
\right),\varphi \left( {\overline x ,\overline y } \right) \ge c +
\varepsilon,-\frac{1}{2} \le x_i  - y_i  \le \frac{1}{2}, i =
\overline{1,n} \} = \emptyset.
\end{array}
\]

Taking into account that $\left| {\varphi \left( {\overline x
,\overline y } \right)} \right| \le 2\sum\limits_{v \in V} {w_v
}$, by the method of binary division we find such a constant $c$,
that the set
\[
\begin{array}{*{20}c}
\Omega = \{\left( {\overline x ,\overline y } \right)|\overline x
\in \Pi \left( {G_1 } \right),\overline y \in \Pi \left( {G_2 }
\right),\varphi \left( {\overline x ,\overline y } \right) \ge
c,-\frac{1}{2} \le x_i  - y_i  \le \frac{1}{2}, i =
\overline{1,n}\} \ne \emptyset
\end{array}
\]
and
\[
\begin{array}{*{20}c}
\{ \left( {\overline x ,\overline y } \right)|\overline x \in \Pi
\left( {G_1 } \right),\overline y  \in \Pi \left( {G_2 }
\right),\varphi \left( {\overline x ,\overline y } \right) \ge c +
\varepsilon ,-\frac{1}{2} \le x_i  - y_i  \le \frac{1}{2}, i =
\overline{1,n} \} = \emptyset.
\end{array}
\]
From lemma we see that
$\overline {\xi ^{opt} }\in \Omega$ and $\varphi(\overline {\xi
^{opt} }) < c + \varepsilon$. And every pair from $\Omega $ is an
$\varepsilon $-solution of the task. Theorem proved.

Consider the following approximate algorithm for 2-MaxCMS.

 1. Find a pair $\left(
{\overline {x'} ,\overline {y'} } \right)$ such that $\mathop
{\max }\limits_{\left( {\overline x ,\overline y } \right) \in \Pi
\left( {G_1 } \right) \times \Pi \left( {G_2 } \right)} \varphi
\left( {\overline x ,\overline y } \right)\le\varphi \left(
{\overline {x'} ,\overline {y'} } \right) + \varepsilon$ и $\left|
{x'_i - y'_i } \right| \le \frac{1}{2}$ where $\varepsilon =
\frac{1}{16}$.

2. Find $\overline x ^*  = \arg \mathop {\max }\limits_{\overline
x \in \Pi \left( {G_1 } \right)} \psi \left( {\overline x
,\overline y ^\prime  } \right)$ and $\overline y ^*  = \arg
\mathop {\max }\limits_{\overline y  \in \Pi \left( {G_2 }
\right)} \psi \left( {\overline x ^* ,\overline y } \right) $.
There $\overline x ^* ,\overline y ^*$ are boolean.

The answer of an algorithm is the set of vertexes $\left\{
{v|x_v^* y_v^* = 1} \right\}$.

It is easy to see that all stages of the algorithm are polynomial.
Let us investigate its answer.

Denote $W = \sum\limits_{v \in V} {w_v }$ and $\varphi \left(
{\overline {x'} ,\overline {y'} } \right) = \alpha W$. It is clear
that $0 \le \alpha \le 1$.

{\bf Theorem 10.} The following is true
\[
\begin{array}{*{20}c}
\mathop {\max }\limits_{\left( {\overline x ,\overline y } \right)
\in \Pi \left( {G_1 } \right) \times \Pi \left( {G_2 } \right)}
\psi \left( {\overline x ,\overline y } \right) - \psi \left(
{\overline {x^* } ,\overline {y^* } } \right) \le \left(
{\frac{1}{4} - \left( {\alpha  - \frac{1}{2}} \right)^2 } \right)W
+ \varepsilon,
\end{array}
\]
if $\alpha \ge \frac{1}{2}$. And also
\[
\mathop {\max }\limits_{\left( {\overline x ,\overline y } \right)
\in \Pi \left( {G_1 } \right) \times \Pi \left( {G_2 } \right)}
\psi \left( {\overline x ,\overline y } \right) - \psi \left(
{\overline {x^* } ,\overline {y^* } } \right) \le \frac{1}{4}W +
\varepsilon,
\]
when $\frac{3}{8} \le \alpha  \le \frac{1}{2}$. And
\[
\begin{array}{*{20}c}
\mathop {\max }\limits_{\left( {\overline x ,\overline y } \right)
\in \Pi \left( {G_1 } \right) \times \Pi \left( {G_2 } \right)}
\psi \left( {\overline x ,\overline y } \right) - \psi \left(
{\overline {x^* } ,\overline {y^* } } \right) \le\left(
{\frac{1}{4} - \left( {\alpha  - \frac{3}{8}} \right)^2 } \right)W
+ \varepsilon,
\end{array}
\]
if $\alpha  \le \frac{3}{8}$.

{\bf Proof.} Let us bound $\varphi \left( {\overline {x'}
,\overline {y'} } \right) - \psi \left( {\overline {x'} ,\overline
{y'} } \right)$, using the fact of concavity of $f\left( x \right)
= x - x^2 $:
\[
\begin{array}{l}
\varphi \left( {\overline {x'} ,\overline {y'} } \right) - \psi \left( {\overline {x'} ,\overline {y'} } \right) = \sum\limits_{v \in V} {\frac{1}{2}w_v \left( {x'_v  - x_v^{'2} } \right) + \frac{1}{2}w_v \left( {y'_v  - y_v^{'2} } \right)}  \le \\
\le \sum\limits_{v \in V} {w_v \frac{{\left( {x'_v  + y'_v } \right)}}{2}\left( {1 - \frac{{\left( {x'_v  + y'_v } \right)}}{2}} \right)}  = \sum\limits_{v \in V} {w_v \left( {\frac{1}{4} - \left( {\frac{{x'_v  + y'_v  - 1}}{2}} \right)^2 } \right)}\\
\end{array}
\]

When $\alpha  \ge \frac{1}{2}$:
\[
\begin{array}{l}
\alpha W = \varphi \left( {\overline {x'} ,\overline {y'} }
\right) = \sum\limits_{v \in V} { - \frac{1}{2}w_v \left( {x'_v  -
y'_v } \right)^2 }  + \frac{1}{2}w_v y'_v  + \frac{1}{2}w_v x'
\le\sum\limits_{v \in V} {\frac{1}{2}w_v y'_v + \frac{1}{2}w_v x'}
\end{array}
\]
and from this:
\[
\sum\limits_{v \in V} {w_v \frac{{\left( {x'_v  + y'_v  - 1}
\right)}}{2}}  \ge \left( {\alpha  - \frac{1}{2}} \right)W.
\]
Then
\[
\begin{array}{l}
\varphi \left( {\overline {x'} ,\overline {y'} } \right) - \psi
\left( {\overline {x'} ,\overline {y'} } \right) \le
\sum\limits_{v \in V} {w_v \left( {\frac{1}{4} - \left(
{\frac{{x'_v  + y'_v  - 1}}{2}} \right)^2 } \right)} \le
\frac{1}{4}W - t,
\end{array}
\]
where $t = \mathop {\min }\limits_{\sum\limits_{v \in V} {w_v t_v
} \ge \left( {\alpha  - \frac{1}{2}} \right)W} \sum\limits_{v \in
V} {w_v t_v^2 }$. It is obvious that $t = \left( {\alpha  -
\frac{1}{2}} \right)^2 W$. So, we obtain
\[
\varphi \left( {\overline {x'} ,\overline {y'} } \right) - \psi
\left( {\overline {x'} ,\overline {y'} } \right) \le \left(
{\frac{1}{4} - \left( {\alpha  - \frac{1}{2}} \right)^2 }
\right)W.
\]
Then using $\varphi \left( {\overline {x'} ,\overline {y'} }
\right) \ge \mathop {\max }\limits_{\left( {\overline x ,\overline
y } \right) \in \Pi \left( {G_1 } \right) \times \Pi \left( {G_2 }
\right)} \psi \left( {\overline x ,\overline y } \right) -
\varepsilon$ and $\psi \left( {\overline {x^* } ,\overline {y^* }
} \right) \ge \psi \left( {\overline {x'} ,\overline {y'} }
\right)$ we finally obtain:
\[
\begin{array}{l}
\mathop {\max }\limits_{\left( {\overline x ,\overline y } \right)
\in \Pi \left( {G_1 } \right) \times \Pi \left( {G_2 } \right)}
\psi \left( {\overline x ,\overline y } \right) - \psi \left(
{\overline {x^* } ,\overline {y^* } } \right) \le \left(
{\frac{1}{4} - \left( {\alpha  - \frac{1}{2}} \right)^2 } \right)W
+ \varepsilon.
\end{array}
\]

Almost analogous, when $\alpha \le \frac{3}{8}$,
\[
\begin{array}{l}
\alpha W = \varphi \left( {\overline {x'} ,\overline {y'} }
\right) = \sum\limits_{v \in V} { - \frac{1}{2}w_v \left( {x'_v  -
y'_v } \right)^2 }  + \frac{1}{2}w_v y'_v  + \frac{1}{2}w_v
x'\ge\\
\ge\sum\limits_{v \in V} {\frac{1}{2}w_v y'_v + \frac{1}{2}w_v x'}
- \frac{1}{8}W
\end{array}
\]
and from this:
\[
\sum\limits_{v \in V} {w_v \frac{{\left( {x'_v  + y'_v  - 1}
\right)}}{2}}  \le \left( {\alpha  - \frac{3}{8}} \right)W.
\]
Analogously,
\[
\varphi \left( {\overline {x'} ,\overline {y'} } \right) - \psi
\left( {\overline {x'} ,\overline {y'} } \right) \le \frac{1}{4}W
- s,
\]
where  $s = \mathop {\min }\limits_{\sum\limits_{v \in V} {w_v t_v
} \le \left( {\alpha  - \frac{3}{8}} \right)W} \sum\limits_{v \in
V} {w_v t_v^2 }  = \left( {\alpha  - \frac{3}{8}} \right)^2 W$.
And finally,
\[
\begin{array}{l}
\mathop {\max }\limits_{\left( {\overline x ,\overline y } \right)
\in \Pi \left( {G_1 } \right) \times \Pi \left( {G_2 } \right)}
\psi \left( {\overline x ,\overline y } \right) - \psi \left(
{\overline {x^* } ,\overline {y^* } } \right) \le \varphi \left(
{\overline {x'} ,\overline {y'} } \right) - \psi \left( {\overline
{x'} ,\overline {y'} } \right) + \varepsilon \le \\
\le\left( {\frac{1}{4} - \left( {\alpha  - \frac{3}{8}} \right)^2
} \right)W + \varepsilon.
\end{array}
\]
The statement of the theorem in case when $\frac{3}{8} \le \alpha
\le \frac{1}{2}$ is obvious. The theorem proved.

\section{Conclusion}
As mentioned above, MaxCMS can be considered as a subcase for the
vertex cover problem. From this point of view the task of finding
MaxCMS is equivalent to the task of removing "noisy" objects from
the training set with a minimal total weight. Let us compare the
approximation ratio of our algorithm with a well-known, standard
2-approximation of vertex cover, that can be found for any graph
with weighted vertexes in polynomial time\cite{hochbaum}.

It is clear that $\varepsilon$ can be made arbitrarily small and
it does not play any role in the bound of theorem 10 because the
bounded value is integer. So, for simplicity, we will believe that
$\varepsilon = 0$. Let us denote $\varphi \left( {\overline {x'}
,\overline {y'} } \right) = \alpha W \ge W - \Delta = \mathop
{\max }\limits_{\left( {\overline x ,\overline y } \right) \in \Pi
\left( {G_1 } \right) \times \Pi \left( {G_2 } \right)} \psi
\left( {\overline x ,\overline y } \right)$.

It is obvious that the ratio 2 of approximation has a meaning only
if maximal consistent with monotonicity set of a special orgraph
has a weight more than half of the sum of weights of all vertexes,
i.e. $\alpha \ge\alpha'= \frac{MaxCMS}{W} \ge \frac{1}{2}$. In
this case from theorem 10 we obtain that:
\[
\mathop {\max }\limits_{\left( {\overline x ,\overline y } \right)
\in \Pi \left( {G_1 } \right) \times \Pi \left( {G_2 } \right)}
\psi \left( {\overline x ,\overline y } \right) - \psi \left(
{\overline {x^* } ,\overline {y^* } } \right) \le\alpha(1-\alpha)W
\le \alpha'(1-\alpha')W =\alpha'\Delta
\]
which means that our algorithm has an approximation ratio equal to
$1+\alpha' \le 2$.

For "almost correct" data, i.e. when $\alpha' \approx 1$,
algorithm has an approximation ratio close to standard 2. But for
"noisy" data it appears to be better than standard. For extreme
case when $\alpha' \approx \frac{1}{2}$ standard 2-approximation
means there is no guarantee that we will not remove all objects as
"noise". On the contrary, the total weight of objects removed by
our algorithm in any case can not exceed optimal solution by more
than $\frac{1}{4}W$. And the bound of theorem 10 shows that our
algorithm can find good approximations to MaxCMS for cases when
even more than half of the training set consists of "noisy"
data($\alpha \le \frac{3}{8}$).





%

\end{document}